# Deep Belief Network Training Improvement Using Elite Samples Minimizing Free Energy

Mohammad Ali Keyvanrad [a], Mohammad Mehdi Homayounpour [a]

[a] *Laboratory for Intelligent Multimedia Processing (LIMP), Computer Engineering and Information Technology Department, Amirkabir University of Technology, Tehran, Iran*

Phone: +98(21)64542747

Fax: +98(21)64542700

Email: {keyvanrad, homayoun}@aut.ac.ir

Abstract. Nowadays this is very popular to use deep architectures in machine learning. Deep Belief Networks (DBNs) are deep architectures that use stack of Restricted Boltzmann Machines (RBM) to create a powerful generative model using training data. In this paper we present an improvement in a common method that is usually used in training of RBMs. The new method uses free energy as a criterion to obtain elite samples from generative model. We argue that these samples can more accurately compute gradient of log probability of training data. According to the results, an error rate of 0.99% was achieved on MNIST test set. This result shows that the proposed method outperforms the method presented in the first paper introducing DBN (1.25% error rate) and general classification methods such as SVM (1.4% error rate) and KNN (with 1.6% error rate). In another test using ISOLET dataset, letter classification error dropped to 3.59% compared to 5.59% error rate achieved in those papers using this dataset. The implemented method is available online at "http://ceit.aut.ac.ir/~keyvanrad/DeeBNet Toolbox.html".

## 1. Introduction

Since many years ago, artificial neural networks have been used in artificial intelligence applications. Pattern recognition, voice and speech analysis and natural language processing are some of these applications that use artificial neural networks. Due to some theoretical and biological reasons, deep models and architectures with many nonlinear processing layers were suggested.

These deep models have many layers and parameters that must be learnt. When the learning process is so complicated and a huge number of parameters are needed, artificial neural networks are rarely used. Problem of this number of layers is that training is time consuming and training becomes trapped at local minima. Therefore we can't achieve acceptable results. One important tool for dealing with this problem is to use DBNs (Deep Neural Network) that can create neural networks including many hidden layers [1].

Deep Belief Networks can be used in classification and feature learning. Data representation is very important in machine learning. Therefore much work has been done for feature preprocessing, feature extraction and feature learning. In feature learning, we can create a feature extraction system and then use the extracted features in classification and other applications. Using unlabeled data in high level feature extraction [2] and also increasing discrimination between extracted features are the benefits of DBN for feature learning [3].

Layers of DBN are created from Restricted Boltzmann Machine (RBM) that is a generative and undirected probabilistic model. RBMs use a hidden layer to model the probability distribution of visible variables. Indeed we can create a DBN for hierarchical processing using stacking RBMs. Therefore most of improvements in DBNs are due to improvement in RBMs. This paper studies improvement in computing gradient of log probability of training data to train RBM model.

Hinton presented DBNs and used it in the task of digit recognition on MNIST data set [4]. He used a DBN with 784-500-500-2000-10 structure, where the first layer possesses 784 features from 28*28 MNIST digit images. The last layer is related to 10 digit labels and other three layers are hidden layers with stochastic binary neurons. Finally this paper achieved 1.25% classification error rate on MNIST test data set.

In another paper from this author [3], he used DBN as a nonlinear model for feature extraction and dimension reduction. Indeed the DBN may be considered as a model that can generate features in its last layer with the ability to reconstruct visible data from generated features. When a general Neural Network is used with many layers, the Neural Network becomes trapped in local minima and the performance will decrease. Therefore determining the initial values for NN weights is critical.

Another paper proposed DDBN (Discriminative Deep Belief Network) is based on DBN as a new classifier [1]. This paper showed the power of DBN in using unlabeled data and also performance improvement by increasing layers (even by 50 hidden layers).

DBN applications are not limited to image processing and can be used in voice processing [5]–[8] with significant efficiency. Most of RBM improvements in this paper are related to model learning. Also the idea of this paper is improvement in computing the gradient of log probability of training data. In this new method, elite samples are obtained from DBN model using free energy, so gradient will be computed more accurately. According to the results, performance and training runtime are comparable with other sampling methods such as CD and PCD.

The rest of this paper is organized as follows: in section 2, RBM and DBN are described. FEPCD (Free Energy in Persistent Contrastive Divergence) that is the proposed method in this paper is presented in section 3. In section 4, some experiments are conducted and the proposed method is compared to some other methods such as CD and PCD in the tasks of digit recognition on MNIST data set and prediction of which letter-name was spoken on ISOLET dataset. Finally, section 5 concludes the paper.

## 2. Deep Belief Networks (DBNs) and Restricted Boltzmann Machines (RBMs)

DBNs are composed of multiple layers of RBMs. RBM is a Boltzmann machine where the connections between hidden visible layers are disjointed. Also the Boltzmann machine is an undirected graphical model (or Markov Random Field). In the Following section, the RBMs and some revised version of RBMs are discussed. It is explained how DBNs are constructed using Restricted Boltzmann Machines (RBMs).

The Boltzmann Machine is a type of MRF. The Boltzmann Machine is a concurrent network with stochastic binary units. The network has a set of visible units $v \in \{0,1\}^{g_v}$ and a set of hidden units $h \in \{0,1\}^{g_h}$ where $g_v$ and $g_h$ are the number of visible units and the number of hidden units respectively (left figure in Figure 1). The energy of the joint configuration $\{v, h\}$ in Boltzmann machine is given as follows:

$$E(v, h) = -\frac{1}{2}v^T L v - \frac{1}{2}h^T J h - v^T W h \tag{1}$$

The bias is removed for simplicity of presentation. The term $W$ is the concurrent weights between visible and hidden units, $L$ is the concurrent weights between visible and visible units and finally $J$ is the concurrent weights between hidden and hidden units. Diagonal values of $L$ and $J$ are zero.

Since Boltzmann machines have a complicated theory and formulations, therefore Restricted Boltzmann Machines are used for simplicity. If $J = 0$ and $L = 0$, the famous RBM model is introduced (the right hand figure in Figure 1).

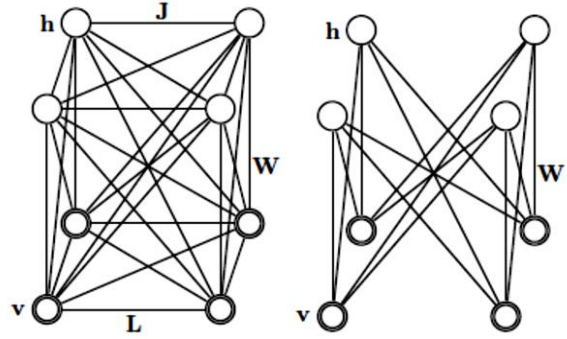

**Figure 1.** Left hand side figure: a general Boltzmann machine. The top layer shows stochastic binary hidden units and the bottom layer shows stochastic binary visible units. Right hand side figure: A restricted Boltzmann machine. the joints between hidden units and also between visible units are disconnected [9].

The energy of the joint configuration {v, h} in restricted Boltzmann machine, with respect to adding bias is given by:

$$E(v, h) = -v^T W h - a^T v - b^T h$$
$$= -\sum_{i=1}^{g_v} \sum_{j=1}^{g_h} W_{ij} v_i h_j - \sum_{i=1}^{g_v} a_i v_i - \sum_{j=1}^{g_h} b_j h_j \quad (2)$$

Where $W_{ij}$ represents the symmetric interaction term between visible unit $i$ and hidden unit $j$, while $b_i$ and $a_j$ are bias terms for hidden units and visible units respectively. The network assigns a probability value with energy function to each state in visible and hidden units.

Because potential functions in MRFs are strictly positive, it is convenient to express them as exponential and Boltzmann distribution [10]. The joint distribution is defined as the product of potentials, and so the total energy is obtained by adding the energies for potential functions. Therefore joint probability distribution for visible and hidden units can be defined as:

$$P(v, h) = \frac{1}{Z} \exp(-E(v, h)) \quad (3)$$

Where $Z$ as partition function or normalization constant, is obtained by summing over all possible pairs of visible and hidden vectors.

$$Z = \sum_v \sum_h \exp(-E(v,h)) \tag{4}$$

The probability assigned to a visible vector $v$ by the network, is obtained by marginalizing out hidden vector $h$.

$$P(v) = \sum_h P(v,h) = \frac{1}{Z} \sum_h \exp(-E(v,h)) \tag{5}$$

The probability that the network assigns to a training image can be increased by adjusting the weights and biases to lower the energy of that image and to raise the energy of other images, especially those images that have low energies and therefore make a big contribution to the partition function [11]. Therefore, best value for each parameter can be found using the following objective function:

$$maximize_{\{w_{ij}, a_i, b_j\}} \frac{1}{m} \sum_{l=1}^m \log\left(\sum_h P(\boldsymbol{v}^{(l)}, \boldsymbol{h}^{(l)})\right) \tag{6}$$

Where the parameter $m$ is the number of training data samples and the aim is to increase the model probability for these training data. Therefore the partial derivative with respect to $w_{ij}$ of the above objective is given by [12] :

$$\begin{aligned}\frac{\partial}{\partial w_{ij}} &\left(\frac{1}{m} \sum_{l=1}^m \log\left(\sum_h P(\boldsymbol{v}^{(l)}, \boldsymbol{h}^{(l)})\right)\right) \\ &= \frac{1}{m} \sum_{l=1}^m \sum_h X_{il} h_j P(h|v=x) \\ &\quad - \sum_{v'} \sum_{h'} v'_i h'_j P(v', h')\end{aligned} \tag{7}$$

Where $X_{il}$ refers to the $i^{th}$ unit of the $l^{th}$ data instance. The sum on the left hand side can be computed exactly; however the expectation on the right hand side (also called the expectation under the model distribution) is intractable. Therefore other methods are used to estimate this partial derivative. The derivative of the log probability of a training vector with respect to a weight can be computed as follows:

$$-\frac{\partial \log P(v)}{\partial w_{ij}} = <v_i h_j>_{data} - <v_i h_j>_{model} \qquad (8)$$

Where the angle brackets are used to denote expectations under the distribution specified by the subscript that follows. This leads to a very simple learning rule for performing stochastic steepest ascent in the log probability of the training data:

$$\Delta w_{ij} = \epsilon \left( <v_i h_j>_{data} - <v_i h_j>_{model} \right) \qquad (9)$$

Where $\epsilon$ parameter is a learning rate. Similarly the learning rule for the bias parameters is:

$$\Delta a_i = \epsilon \left( <v_i>_{data} - <v_i>_{model} \right) \qquad (10)$$

$$\Delta b_j = \epsilon \left( <h_j>_{data} - <h_j>_{model} \right) \qquad (11)$$

Since there are no direct connections between hidden units in an RBM, these hidden units are independent given visible units [11]. This fact is based on MRF properties [10]. Now Given a randomly selected training image $v$, the binary state $h_j$ of each hidden unit $j$, is set to 1 where its probability is:

$$P(h_j = 1|\boldsymbol{v}) = g\left( b_j + \sum_i v_i w_{ij} \right) \qquad (12)$$

Where $g(x)$ is the logistic sigmoid function $g(x) = 1/(1 + \exp(-x))$. Therefore $<v_i h_j>_{data}$ can be computed easily.

Since there are no direct connections between visible units in an RBM, it is very easy to obtain an unbiased sample of the state of a visible unit, given a hidden vector

$$P(v_i = 1|\boldsymbol{h}) = g\left( a_i + \sum_j h_j w_{ij} \right) \qquad (13)$$

However computing $<v_i h_j>_{model}$ is so difficult. It can be done by starting from any random state of the visible units and performing sequential Gibbs sampling for a long time. Finally due to impossibility of this method and large run-times, Contrastive Divergence (CD) method is used [13].

RBM has many benefits and has been greatly used in recent years, especially in DBN's. Nowadays many papers wish to improve this model and its performance. In the following section these improvements on computing gradient of log probability of train data are discussed.

## 2.1. Computing gradient of log probability of train data

According to equation (5), the $\log P(v)$ can be expressed as follows [14]:

$$\phi = \log P(v) = \phi^+ - \phi^-$$

$$\phi^+ = \log \sum_h \exp(-E(v,h))$$

$$\phi^- = \log Z = \log \sum_v \sum_h \exp(-E(v,h))$$

(14)

The gradient of $\phi^+$ according to model parameters is a positive gradient and similarly, the gradient of $\phi^-$ according to model parameters is a negative gradient.

$$\frac{\partial \phi^+}{\partial w_{ij}} = v_i . P(h_j = 1|v)$$

$$\frac{\partial \phi^-}{\partial w_{ij}} = P(v_i = 1, h_j = 1)$$

(15)

Computing the positive gradient is simple but computing the negative gradient is intractable and therefore inference methods using sampling are used to compute gradient.

Based on the above sections, the gradient of log probability of training data is obtained from equation (8). We must compute $<v_i h_j>_{data}$ and $<v_i h_j>_{model}$ for computing gradient and adjusting parameters according to equation (9). Based on most of the literatures on RBMs, computing $<v_i h_j>_{data}$ is called positive phase, and computing $<v_i h_j>_{model}$ is called negative phase corresponding to positive gradient and negative gradient respectively.

Since there is no interconnections between hidden units and they are independent, $< v_i h_j >_{data}$ can easily be computed by considering the visible units $v$ (that their values have been determined by training data) and assigning the value 1 to each hidden unit with the probability of $P(h_j = 1|v)$ regarding to equation *(12)*.

The main problem resides in the negative phase. In practice, the difference between different DBN learning methods (e.g. Contrastive Divergence or Persistent Contrastive Divergence) is in sampling in their negative phase [15].

To compute $< v_i h_j >_{model}$, Gibbs sampling method may be used. This method starts with random values in visible units and Gibbs sampling steps should continue for a long time. Each Gibbs sampling step leads to updating of all hidden units according to equation *(12)* and then updating all visible units according to equation *(13)* (see Figure 2). Indeed, Gibbs sampling is a method for obtaining a good sample from joint distribution on $v$ and $h$ in this model.

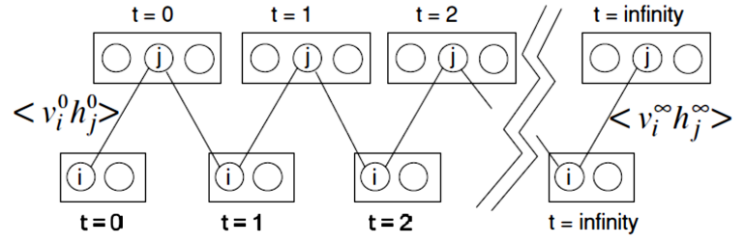

*Figure 2: Gibbs sampling. Each Gibbs sampling step means updating of all hidden units according to equation (12) and then updating all visible units according to equation (13). The chain is initialized by setting the binary states of the visible units to be the same as a data vector [4].*

### 2.1.1. Contrastive Divergence (CD)

Since Gibbs sampling method is slow, Contrastive Divergence (CD) algorithm is used [13]. In this method visible units are initialized using training data. Then binary hidden units are computed according to equation *(12)*. After determining binary hidden unit states, $v_i$ values are recomputed according to equation *(13)*. Finally, probability of hidden unit activations is computed and using these values of hidden units and visible units, $< v_i h_j >_{model}$ is computed. The computation steps in $CD_1$ method is graphically illustrated in Figure 3.

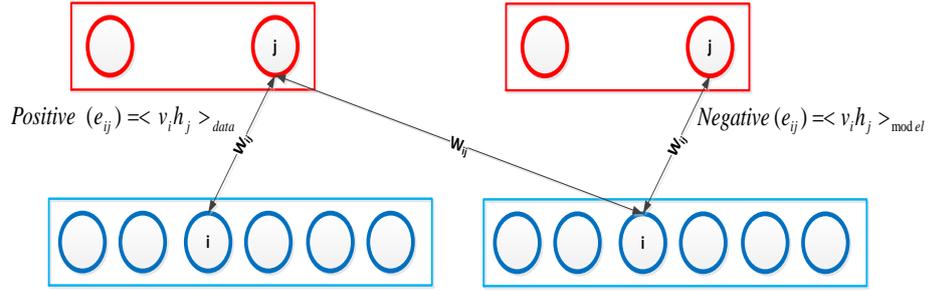

*Figure 3: Computation steps in $CD_1$ method. $Positive\ (e_{ij})$ is related to computing $<v_i h_j>_{data}$ for $e_{ij}$ connection.*

Although $CD_1$ method is not a perfect gradient computation method, but its results are acceptable [13]. By repeating Gibbs sampling steps, $CD_k$ method is achieved. The k parameter is the number of repetitions of Gibbs sampling steps. This method has a higher performance and can compute gradient more exactly [16].

*2.1.2. Persistent Contrastive Divergence (PCD)*

Whereas $CD_k$ has some disadvantages and is not exact, other methods are proposed in RBM. One of these methods is PCD that is very popular [17]. Unlike CD method that uses training data as initial value for visible units, PCD method uses last chain state in the last update step. In other words, PCD uses successive Gibbs sampling runs to estimate $<v_i h_j>_{model}$. Although all model parameters are changed in each step, but can receive good samples from model distribution with a few Gibbs sampling steps because the model parameters change slightly [18]. Many persistent chains can be run in parallel and we will refer to the current state in each of these chains as new sample or a "fantasy" particle [9], [17]. Improvement in PCD method is the novelty of this paper that will be described in the section 3.

## 2.2. Deep Belief Network

After an RBM has been learned, the activities of its hidden units (when they are being driven by data) can be used as the 'data' for learning a higher-level RBM [19]. The idea behind DBN is to allow each RBM model in the sequence to receive a different representation of the data. The model performs a nonlinear transformation on its input vectors and produces as output, the vectors that will be used as input for the next model in the sequence [4].

After layer-by-layer pre-training in DBN, we use back-propagation technique through the whole classifier to fine-tune the weights for optimal classification. Pretraining helps generalization and the very limited information in the data is used only to slightly adjust the weights found by pretraining [3].

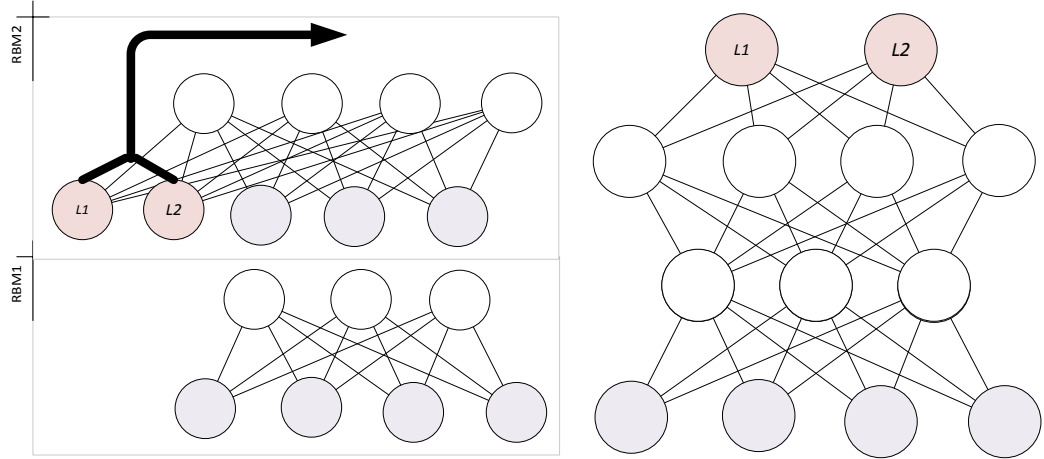

Figure 4: Converting a DBN to a classifier neural network for running back-propagation. Left figure: a DBN with a discriminative RBM in top layer. Right figure: A neural network that is ready for back-propagation.

# 3. Free Energy in Persistent Contrastive Divergence (FEPCD)

One of the main challenges in RBMs is training of their parameters. As discussed before, computing gradient of model is intractable; therefore sampling methods are used for gradient estimation. Sampling methods are used because gradient estimation needs samples from the model that has been trained. Since in an RBM each unit in a layer is independent from other units in other layers, therefore Gibbs sampling is a proper method. But in order to obtain appropriate samples from the model, Gibbs sampling needs to be run for many times and this is impossible. Therefore different methods as CD or PCD have been proposed. In this paper a new method for generating elite samples as described later has been proposed.

In PCD method, as described before, many persistent chains can be run in parallel and we will refer to the current state in each of these chains as a "fantasy" particle. Chain selection in this method is blind and the best one may not be selected. If we can define a criterion for goodness of a chain, samples and therefore computing gradient will be more accurate.

The proposed criterion for selecting the best chain is the free energy of visible sample $v$ which is defined as follows [11]:

$$P(v) = \frac{1}{Z} e^{-F(v)} = \frac{1}{Z} \sum_h e^{-E(v,h)} \tag{16}$$

where $F(v)$ is free energy. Therefore $F(v)$ can be computed as follows [11]:

$$F(v) = -\sum_i v_i a_i - \sum_j q_j I_j + \sum_j (q_j \log q_j + (1-q_j) \log(1-q_j)) \tag{17}$$

Where $I_j = b_j + \sum_i v_i w_{ij}$ is equal to sum of inputs to hidden unit $j$ and $q_j = g(I_j)$ is equal to activation probability of hidden unit $h_j$ given $v$ and $g$ is logistic function. An equivalent and simpler equation for computing $F(v)$ is as follows:

$$F(v) = -\sum_i v_i a_i - \sum_j \log(1 + e^{I_j}) \tag{18}$$

To understand the criterion benefit, we must describe the training phase in more details. According to equation (3) and (5), $P(v)$ can be defined as follows:

$$P(v,h) = \frac{e^{-E(v,h)}}{\sum_{v',h'} e^{-E(v',h')}} \rightarrow P(v) = \frac{\sum_h e^{-E(v,h)}}{\sum_{v',h'} e^{-E(v',h')}} \tag{19}$$

Thus the derivative of $P(v)$ according to any parameter $\theta$ is as follows [16]:

$$\begin{aligned}
\frac{\partial \log P(v)}{\partial \theta} &= \frac{\partial \log \sum_h e^{-E(v,h)}}{\partial \theta} - \frac{\partial \log \sum_{v',h'} e^{-E(v',h')}}{\partial \theta} \\
&= -\frac{1}{\sum_h e^{-E(v,h)}} \sum_h e^{-E(v,h)} \frac{\partial E(v,h)}{\partial \theta} \\
&\quad + \frac{1}{\sum_{v',h'} e^{-E(v',h')}} \sum_{v',h'} e^{-E(v',h')} \frac{\partial E(v',h')}{\partial \theta} \\
&= -\sum_h P(h|v) \frac{\partial E(v,h)}{\partial \theta} \\
&\quad + \sum_{v',h'} P(v',h') \frac{\partial E(v',h')}{\partial \theta}
\end{aligned} \tag{20}$$

According to equation (*2*), the energy function is very simple and $\frac{\partial E(v,h)}{\partial \theta}$ can be easily computed for any parameter $\theta$. In general, calculating expectation of arbitrary function $f$ ( $\mathbb{E}(f) = \sum_z P(z) f(z)$ ) can be computed using sampling methods [10] (sampling $z$ from $P(z)$ and average on $f(z)$, $\mathbb{E}(f) = \frac{1}{n}\sum_{i=1}^{n} f(z)$). So in equation (*20*), we need to sample from $P(h|v)$ and $P(v', h')$. In sampling for $P(h|v)$, $v$ is clamped to the visible input vector, and we sample $h$ given $x$. But sampling for $P(v', h')$ is more difficult and both $v$ and $h$ are sampled. Due to this difficulty, Gibbs sampling method is used. Based on earlier description, Gibbs sampling is not practicable and other methods like CD or PCD are used instead. In these methods, sampling runs a few steps and therefore samples do not correspond to the model distribution exactly. Therefore in our FEPCD method, we try to find those better chains in PCD method that have more similarity to model distribution (or have greater $P(v)$ ). By using these selected samples, gradient of $P(v)$ will be more accurate and the error of samples obtained from the model will reduce.

Finally we need a criterion that depicts the sampling chain goodness. This criterion can be $P(v_{gen})$ where $v_{gen}$ are the samples generated from the model using sampling chains. Samples with greater $P(v_{gen})$ are then selected as elite samples obtained from the model. But computing $P(v_{gen})$ is intractable and therefore another criterion was proposed. Since the parameters of model are fixed for all generated samples, according to equation (*16*), the partition function Z is the same for these samples and the $P(v_{gen})$ is only related to $F(v_{gen})$. Therefore the generated samples with lower $F(v_{gen})$ are the best model samples in equation (*20*).

$$\begin{aligned}\left\{v_{gen} \in generated\ samples : P(v_{gen}) = \frac{1}{Z} e^{-F(v_{gen})} > \delta\right\} \\ = \left\{v_{gen} \in generated\ samples : F(v_{gen}) < \delta'\right\}\end{aligned} \quad (21)$$

where $\delta'$ is the threshold to determine good samples. In our experiments we select half of generated data with lower $F(v_{gen})$ as the elite samples and therefore computing the threshold is not necessary.

# 4. Results

The method proposed in this paper was evaluated by applying it to the MNIST and ISOLET dataset. Also we used the DeeBNet toolbox [20] (the implemented toolbox with authors) in the experiments. In addition, our FEPCD method has been implemented and is available online[1].

## 4.1. MNIST dataset

MNIST dataset includes images of handwritten digits [21] (10 classes of digits 0-9). Each digit was cared to be located in the center of each 28*28 image. The image pixels have discrete values between 0 and 255 that most of them have the values at the edge of this interval [22]. The image pixel values were normalized between 0 and 1. The dataset was divided to train and test parts including 60,000 and 10,000 images respectively[2]. In our experiments, these discrete values have been mapped to interval [0-1] using min-max normalization method.

In the first experiment a discriminative RBM has been used. Structure of this RBM is 784-500. In other words this RBM has 784 visible units (images has 28*28 pixels) and 500 binary hidden units. Classification is done by computing $P(y|v)$ in each class [11], [23]. The results are presented in Table 1. The results have been obtained using 10 separate runs on MNIST.

Table 1: RBM classification error for digit recognition on MNIST dataset using different sampling methods in training phase. The results were achieved using 10 separate runs.

| Method | Error  | Variance |
|--------|--------|----------|
| CD     | 0.1079 | 0.00504  |
| PCD    | 0.0340 | 0.00112  |
| FEPCD  | 0.0318 | 0.00039  |

According to Table 1, the proposed FEPCD sampling method is more appropriate relating to other sampling methods such as CD or PCD. These results show that gradient is computed using better and more accurate samples.

Another aspect to be considered is the training speed in each method. Naturally the FEPCD is more slowly than other methods since in this method we must compute free energy that is partly time consuming. Since training is offline, so a long training time may be acceptable, but according to Figure 5 when PCD gets to its best results (after second 500), the FEPCD method reaches to its best results in similar time with fewer epochs. This figure shows that FEPCD is slower in each epoch, but it converges in fewer epochs, and its training time is close to that of CD and PCD methods.

---
[1] Available online at "http://ceit.aut.ac.ir/~keyvanrad/DeeBNet Toolbox.html"
[2] Available online at "http://yann.lecun.com/exdb/mnist/"

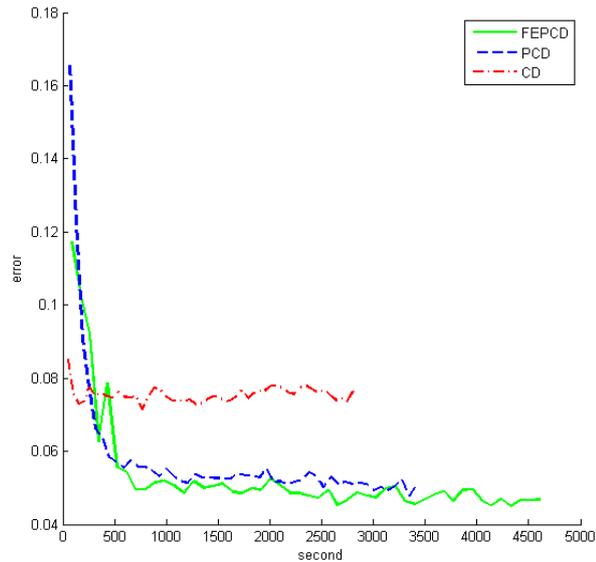

Figure 5: Training speed and efficiency in different sampling methods. Horizontal axis is training time in seconds and vertical axis is classification error on MNIST test set.

Finally in the main experiment, a DBN with an structure of 784-500-500-2000, similar to the structure proposed by Hinton [4] was trained. As described before, in many DBN papers, after training a DBN, it can be fine-tuned using Back-Propagation (BP) method [1], [3]. The results are given in Table 2. In these results, each RBM in DBN (before back-propagation) was trained in 50 epochs. Then the DBN was fine-tuned in 200 epochs using back-propagation method.

Table 2: Classification error on MNIST dataset for a DBN (784-500-500-2000) using different sampling methods. Each RBM in DBN (before back-propagation) is trained in 50 epochs. Then the DBN was fine-tuned in 200 epochs using back-propagation method.

| Method | Before BP | After BP |
|--------|-----------|----------|
| CD     | 0.0636    | 0.0124   |
| PCD    | 0.0307    | 0.0122   |
| FEPCD  | **0.0248** | **0.0111** |

According to Table 2, FEPCD, can improve performance before and after using back-propagation. In another test, each RBM was trained in 200 epochs and the best result was obtained. In this experiment the test error rate decreased to 0.0099. These results are better than the best results obtained in Hinton [4] (0.0125,error rate), and other classification methods such as NN (0.016 error rate) or SVM (0.014 error rate) [3].

### 4.2. ISOLET dataset

In ISOLET data set, 150 subjects utter twice the name of each letter of the alphabet. Hence, there are 52 training examples from each speaker. The speakers are grouped into sets of 30 speakers each, and are referred to as isolet1, isolet2, isolet3, isolet4, and isolet5. The data appears in isolet1+2+3+4 data in a sequential order, i.e. first the speakers from isolet1, then isolet2, and so on. The test set, isolet5, is a separate file[3]. Due to missing three examples, there are 7797 examples in total referred to as isolet1-isolet5 (6238 training examples and 1559 test examples). The features vector has 617 features including spectral coefficients, contour features, sonorant features, pre-sonorant features, and post-sonorant features [24].

Since the features have real values, the Gaussian visible units is used [11]. Similar to MNIST test, in the first experiment a discriminative RBM has been applied. Structure of this RBM is 617-1000. The results are presented in Table 3. The results have been obtained using 10 separate runs on ISOLET dataset. In another test, a DBN with a structure of 617-1000-1000-2000 was trained. As described before, after training the DBN, fine tuning is done using Back-Propagation (BP) method. The results are presented in Table 3. In these results, each RBM in DBN (before back-propagation) was trained in 200 epochs. Then the DBN was fine-tuned in 200 epochs using back-propagation method. In these DBNs, first layer RBM is trained with different sampling methods as first layer feature extractor and the two other layers are trained with CD sampling method.

The conducted experiments show again the capability of the proposed method to obtain more accurate gradients of log probability of training data and achieve smaller classification error. The best result was obtained with FEPCD method with 0.0353 classification error rate that is better than the best results obtained in new articles like [25] with 0.0559 classification error rate.

---

[3] Available online at "https://archive.ics.uci.edu/ml/datasets/ISOLET"

Table 3: RBM classification error for digit recognition on ISOLET dataset using different sampling methods in training phase with 10 separate runs and classification error for a DBN (617-1000-1000-2000) using different sampling methods in first layer. Each RBM in DBN (before back-propagation) is trained in 200 epochs. Then the DBN was fine-tuned in 200 epochs using back-propagation method.

| Method | RBM | | DBN | |
|---|---|---|---|---|
| | Error | Variance | Before BP | After BP |
| **CD** | 0.0890 | 0.000012 | 0.0552 | 0.0372 |
| **PCD** | 0.0843 | 0.000016 | 0.0500 | 0.0385 |
| **FEPCD** | **0.0791** | 0.000020 | **0.0449** | **0.0353** |

# 5. Conclusion

In this paper we discussed one of the main problems in learning Deep Belief Networks. From the beginning of invention of DBN, since the gradient of log probability of training data is intractable, therefore the sampling methods are used to estimate this gradient. The main part of this estimation is sampling from the model that its parameters have been tuned using training data. In our new method, the elite samples are found in multiple sampling chains using free energy value, which is proportional to probability of training data. These selected samples can be used in computation of gradient with more accuracy. According to the results, this method improves performance on MNIST and ISOLET datasets and outperforms other general sampling methods such as CD or PCD.

For future work, we would like to use free energy criterion as a fitness function in evolutionary algorithms. In this new idea, we can achieve new improved chains using evolutionary operations.

Another improving idea is to use both CD and FEPCD sampling methods simultaneously, and to use their both advantages. Although the CD method has low performance, but in the first epochs, CD has better gradient computation. Now, we can improve the training speed of FEPCD by merging it to CD.